\PassOptionsToPackage{obeyspaces}{url}
\documentclass[sigconf,nonacm,screen]{acmart}
\usepackage[acm,subfig]{definition}

\author[Yanqiao Zhu, Yichen Xu, Hejie Cui, Carl Yang, Qiang Liu, and Shu Wu]{Yanqiao Zhu$^{1,2,}$*, Yichen Xu$^{3,}$*, Hejie Cui$^{4}$, Carl Yang$^{4}$, Qiang Liu$^{1,2}$, and Shu Wu$^{1,2,\dagger}$}

\authornotetext{The first two authors made equal contribution to this work.}
\authornotetext{To whom correspondence should be addressed.}

\affiliation{%
	\institution{$^1$Center for Research on Intelligent Perception and Computing, Institute of Automation, Chinese Academy of Sciences}
	\institution{$^2$School of Artificial Intelligence, University of Chinese Academy of Sciences}
	\institution{$^3$School of Computer Science, Beijing University of Posts and Telecommunications}
	\institution{$^4$Department of Computer Science, Emory University}
	\country{}
}

\email{yanqiao.zhu@cripac.ia.ac.cn, linyxus@bupt.edu.cn, {hejie.cui, j.carlyang}@emory.edu, {qiang.liu, shu.wu}@nlpr.ia.ac.cn}

\begin{document}

\newcommand{\themodel}{HORACE\xspace}

\title{Structure-Aware Hard Negative Mining \mbox{for Heterogeneous Graph Contrastive Learning}}

\begin{abstract}
Recently, heterogeneous Graph Neural Networks (GNNs) have become a de facto model for analyzing HGs, while most of them rely on a relative large number of labeled data.
In this work, we investigate Contrastive Learning (CL), a key component in self-supervised approaches, on HGs to alleviate the label scarcity problem.
We first generate multiple semantic views according to metapaths and network schemas. Then, by pushing node embeddings corresponding to different semantic views close to each other (positives) and pulling other embeddings apart (negatives), one can obtain informative representations without human annotations.
However, this CL approach ignores the relative hardness of negative samples, which may lead to suboptimal performance. Considering the complex graph structure and the smoothing nature of GNNs, we propose a structure-aware hard negative mining scheme that measures hardness by structural characteristics for HGs. By synthesizing more negative nodes, we give larger weights to harder negatives with limited computational overhead to further boost the performance.
Empirical studies on three real-world datasets show the effectiveness of our proposed method. The proposed method consistently outperforms existing state-of-the-art methods and notably, even surpasses several supervised counterparts.
\end{abstract}


\maketitle

\section{Introduction}

Many real-world complex interactive objectives can be represented in Heterogeneous Graphs (HGs) or heterogeneous information networks.
Recent development in heterogeneous Graph Neural Networks (GNNs)
has achieved great success in analyzing heterogeneous structure data \cite{Shi:2019kf,Sun:2011be}.
However, most existing models require a relatively large amount of labeled data for proper training \cite{Wang:2019gv,Kipf:2017tc,Velickovic:2019tu,Wang:2019gv,Fu:2020fs}, which may not be accessible in reality.
As a promising strategy of leveraging abundant unlabeled data, Contrastive Learning (CL), as a case of self-supervised learning, is proposed to learn representations by distinguishing semantically similar samples (positives) over dissimilar samples (negatives) in the latent space \cite{Chen:2020wj,Grill:2020uc,Chen:2020uu,Tian:2020vw,Caron:2020uv}.

Most existing CL work follows a multiview paradigm, where multiple views of the input data are constructed via semantic-preserving augmentations.
In the HG domain, since multiple types of nodes and edges convey abundant semantic information, it is straightforward to construct views based on HG semantics such as metapaths and schemas. In this way, for one anchor node, its embeddings in different semantic views constitute positives and all other embeddings are naturally regarded as negative examples.

However, the previous scheme assumes that all negative samples make equal contribution to the CL objective.
Previous research in metric learning \cite{Schroff:2015wo} and visual representation learning \cite{Cai:2020tz,Xuan:2020is} has established that the \emph{hard negative sample} is of particular concern in effective CL. To be specific, the more similar a negative sample to its anchor, the more helpful it is for learning effective representatives.
Therefore, a natural question arises: \emph{whether could we make use of hard negatives in HGCL?} To this end, we propose to investigate the relative hardness of different negative samples in HGs and reweight hard negatives to further boost performance of CL in HGs.

\begin{figure}[t]
	\centering
	\includegraphics[width=0.85\linewidth]{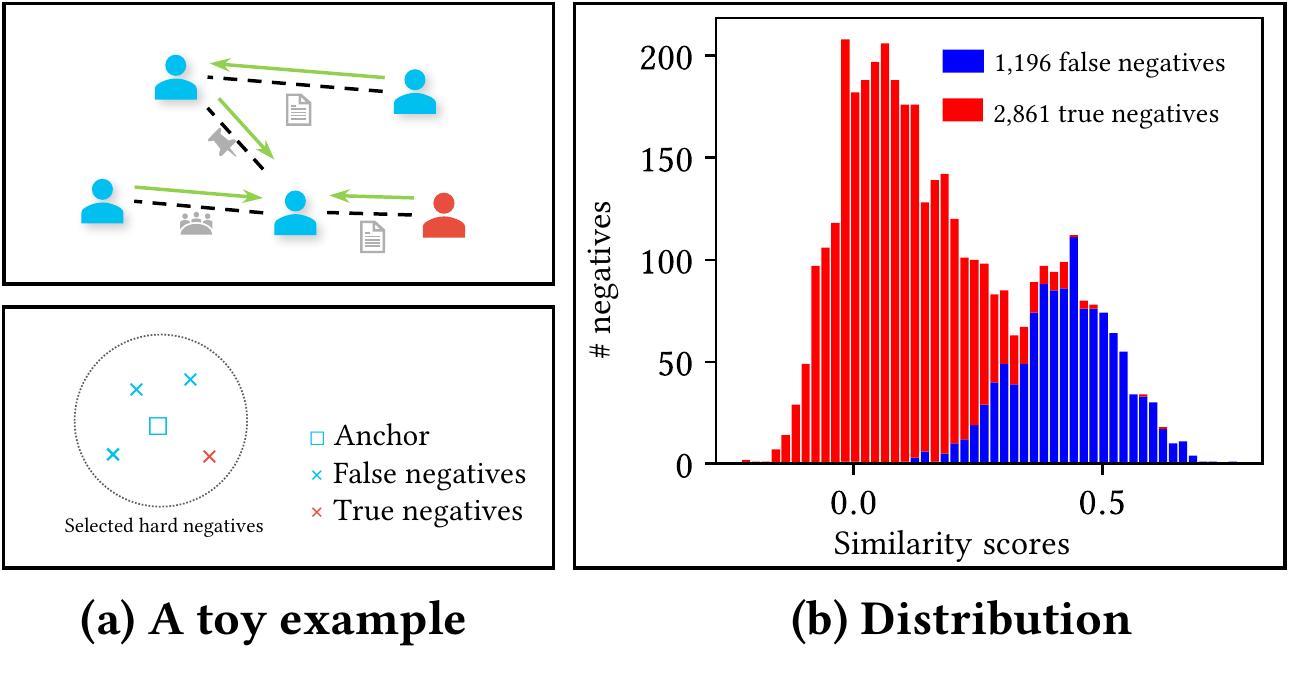}
	\caption{(a) A toy example of an academic network. Heterogeneous GNNs produce similar embeddings for nodes sharing the same label in its ego network by aggregating semantic-specific neighborhood information. (b) A histogram of negatives and their semantic similarity scores with an anchor node. With the similarity to the anchor node increasing, there are more positive samples ({\color{blue}false negatives}), leading to wrong selection of hard negatives.}
	\label{fig:pos-neg-DBLP}
\end{figure}

In visual CL studies, the hardness of one image is defined to be its semantic similarity to the anchor, e.g., inner product of two normalized vectors in the embedding space.
When dealing with HGs, due to the neighborhood aggregation scheme in each semantic view \cite{Wang:2019gv}, heterogeneous GNN produces similar embeddings within ego networks; embeddings of neighboring nodes sharing \emph{the same label} with the anchor node thus tend to be similar to the anchor, as shown in Figure \ref{fig:pos-neg-DBLP}(a).
We further plot pairwise similarities of negative nodes with one arbitrary author node in a real-world DBLP network in Figure \ref{fig:pos-neg-DBLP}(b). With the similarity of negative node to the anchor (the hardness) increasing, there are more \emph{positive} samples (\emph{false} negatives). Therefore, measuring semantic hardness simply by embedding similarities results in \emph{hard but false} negatives being selected, which inevitably impairs the performance.
Furthermore, at the beginning of training, node embeddings are suffered from poor quality, which may be another obstacle of selecting hard negative samples.

\begin{figure}
    \centering
    \includegraphics[width=\linewidth]{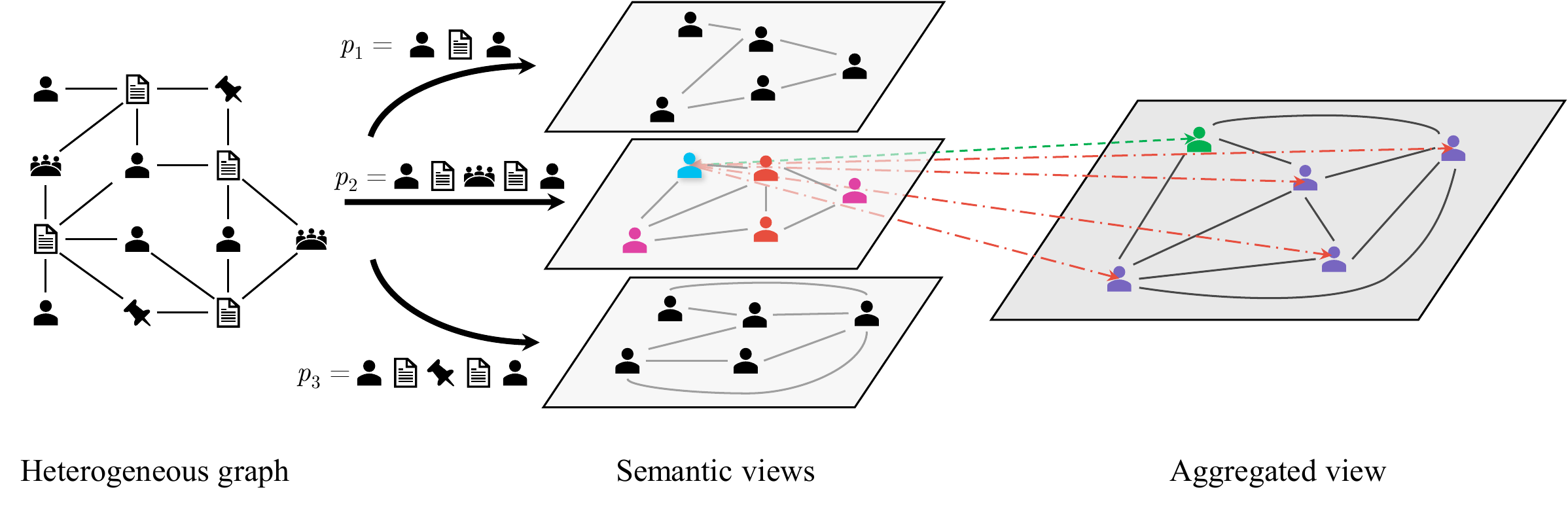}
    \caption{Illustrating the proposed method. We construct semantic views and learn representations with heterogeneous GNNs (\S \ref{sec:preliminaries}). Then, we train the model with a multiview contrastive objective (\S \ref{sec:HGCL}). Take \(\bm{h}_1^{p_2}\) as {\color{sblue} an anchor} for example. Its {\color{sgreen} positive sample} is the aggregated representation \(\bm{h}_1\); {\color{sred} intra-view negatives} \(\bm{h}_2^{p_2}\) to \(\bm{h}_5^{p_2}\) and {\color{spurple} inter-view negatives} \(\bm{h}_2\) to \(\bm{h}_5\) constitute its negatives. Structurally hard negatives discovered by our algorithm are highlighted in {\color{spink} pink} (\S \ref{sec:hard-negative-mining}).}
    \label{fig:framework}
\end{figure}

The above data analysis motivates us to discover hard negatives for HGs from \emph{structural aspects}.
Specifically, we propose to select hard negative nodes according to structural measures and synthesize more negatives by randomly mixing up these selected negatives, so as to give larger weights to harder negatives.
Moreover, measuring hardness through structural characteristics enjoys another benefit that being irrespective of training progress, which can be used in supplement to poor representations in the initial training stage.
We term the resulting CL framework for HGs as \underline{H}eter\underline{O}geneous g\underline{RA}ph \underline{C}ontrastive learning with structure-aware hard n\underline{E}gative mining, \themodel for brevity (Figure \ref{fig:framework}).
The \themodel works by constructing multiple semantic views from the HG at first. Then, we learn node embeddings within each semantic view and combine them into an aggregated representation.
Thereafter, we train the model with a contrastive aggregation objective to adaptively distill information from each semantic view.
Finally, the proposed structure-aware scheme enriches the selection of negatives with structure embeddings, which yields harder negative samples in the context of HGs.

The main contribution of this work is twofold.
\begin{itemize}
	\item We present a CL framework that enables self-supervised training for HGs from both semantic and structural aspects, where we propose to enrich the contrastive objective with structurally hard negatives to improve the performance.
	\item Extensive experiments on three real-word datasets from various domains demonstrate the effectiveness of the proposed method. Particularly, our \themodel method outperforms representative unsupervised baseline methods, achieves competitive performance with supervised counterparts, and even exceeds some of them.
\end{itemize}

\section{The Proposed Method: \themodel}

\subsection{Preliminaries}
\label{sec:preliminaries}

\paragraph{Problem definition.}
We denote a HG with multiple types of nodes and edges by \(\mathcal{G} = (\mathcal{V}, \mathcal{E}, \bm{X}, \phi, \varphi)\), where \(\mathcal{V}, \mathcal{E}\) denote the node set and the edge set respectively. The type mapping function \(\phi\) and \(\varphi\) associates each node and edge with a type respectively.
We further denote the set of all considered metapaths as \(\mathcal{P}\), where a metapath \(p \in \mathcal{P}\) defines a path on the network schema that captures the proximity between the two nodes from a particular semantic perspective.
Given a HG \(\mathcal{G}\), the problem of HG representation learning is to learn node representations \(\bm{H} \in \mathbb{R}^{|\mathcal{V}| \times d}\) that encode both structural and semantic information, where \(d \ll |\mathcal{V}|\) is the hidden dimension.

\paragraph{Heterogeneous graph neural networks.}
Most heterogeneous GNN \cite{Wang:2019gv,Fu:2020fs} learns node representations under different semantic views and then aggregates them using attention networks.
We first generate multiple semantic views, each corresponding to one metapath that encodes one aspect of semantic information. Then, we leverage an attentive network to compute semantic-specific embedding \(\bm{h}_i^p\) for node \(v_i\) under metapath \(p\) as
\begin{equation}
	\bm{h}_i^p = \concat_{k=1}^K \sigma\left( \sum_{v_j \in \mathcal{N}_p(v_i)} \alpha_{ij}^p \bm{W}^p \bm{x}_j \right),
	\label{eq:semantic-view-encoding}
\end{equation}
where \(\parallel\) concatenates \(K\) standalone node representations in each attention head, \(\mathcal{N}_r(v_i)\) defines the neighborhood of \(v_i\) that is connected by metapath \(p\), \(\bm{W}^p \in \mathbb{R}^{d \times m}\) is a linear transformation matrix for metapath \(p\), and \(\sigma(\cdot)\) is the activation function, such as \(\operatorname{ReLU}(\cdot) = \max(0, \cdot)\).
The attention coefficient \(\alpha_{ij}^p\) can be computed by a softmax function
\begin{equation}
	\alpha_{ij}^p = \frac{\exp(\sigma(\bm{a}_p^\top[\bm{h}_i^p \parallel \bm{h}_j^p]))}{\sum_{v_k \in \mathcal{N}_p(v_i)}\exp(\sigma(\bm{a}_p^\top[\bm{h}_i^p \parallel \bm{h}_k^p]))},
\end{equation}
where \(\bm{a}_p \in \mathbb{R}^{2d}\) is a trainable semantic-specific linear weight vector.

Finally, we combine node representation in each view to an aggregated representation.
We employ another attentive network to obtain the semantic-aggregated representation \(\bm{h}_i\) that combines information from every semantic space as given by
\begin{equation}
	\bm{h}_i = \sum_{p = 1}^{|\mathcal{P}|} \beta^p \bm{h}_i^p.\label{eq:semantic-view-aggregation}
\end{equation}
The coefficients are given by
\begin{align}
	\beta^p & = \frac{\exp(w^p)}{\sum_{p' \in \mathcal{P}} \exp(w^{p'})}, \\
	w^p & = \frac{1}{|\mathcal{V}|} \sum_{v_i \in \mathcal{V}} \bm{q}^\top \cdot \tanh(\bm{Wh}_i^p + \bm{b}),
\end{align}
where \(\bm{q} \in \mathbb{R}^{d_m}\) is the semantic-aggregation attention vector, \(\bm{W} \in \mathbb{R}^{d_m \times d}, \bm{b} \in \mathbb{R}^{d_m}\) is the weight matrix and the bias vector respectively, and \(d_m\) is a hyperparameter.

\subsection{Heterogeneous Graph Contrastive Learning}
\label{sec:HGCL}

Existing graph CL follows a multiview framework \cite{Zhu:2020vf,Zhu:2021wh,You:2020ut,Hassani:2020un}, which maximizes the agreement among node representations under different views of the original graph and thus enables the encoder to learn informative representations in a self-supervised manner.

In HGs, since multiple views are involved, to implement the CL objective, we resort to \emph{maximize the agreement between node representations under a specific semantic view and the aggregated representations}.
The resulting contrastive aggregation objective can be mathematically expressed as
\begin{equation}
	\max \frac{1}{|\mathcal{V}|} \sum_{v_i \in \mathcal{V}} \left[ \frac{1}{|\mathcal{P}|} \sum_{p \in \mathcal{P}} \frac{1}{2} \left( I(\bm{h}^p_i; \bm{h}_i) + I(\bm{h}_i; \bm{h}_i^p) \right) \right].
	\label{eq:multiview-contrastive-objective}
\end{equation}
The proposed objective Eq. (\ref{eq:multiview-contrastive-objective}) conceptually relates to contrastive knowledge distillation \cite{Tian:2020tz}, where several teacher models (semantic views) and one student model (the aggregated representation) are employed.
By forcing the embeddings between several teachers and a student to be the same, these aggregated embeddings adaptively collects information of all semantic relations.

Following previous work \cite{Henaff:2020ta,vandenOord:2018ut}, to implement the objective \(I(\bm{h}^p_i; \bm{h}_i)\) in Eq. (\ref{eq:multiview-contrastive-objective}), we empirically choose the InfoNCE estimator. Specifically, for node representation \(\bm{h}^p_i\) in one specific semantic view, we construct its positive sample as the aggregated representation, while embeddings of all other nodes in the semantic and the aggregated embeddings are considered as negatives. The contrastive loss can be expressed by
\begin{equation}
	\ell(\bm{h}^p_i, \bm{h}_i) = - \log\frac{ e^{\theta(\bm{h}^p_{i}, \bm{h}_{i}) / \tau}}
			{e^{\theta(\bm{h}^p_{i}, \bm{h}_{i}) / \tau} + \sum\limits_{j \neq i} \left( e^{\theta(\bm{h}^p_{i}, \bm{h}_{j}) / \tau} + e^{\theta(\bm{h}^p_{i}, \bm{h}^p_{j}) / \tau} \right) },
	\label{eq:semantic-contrastive-loss}
\end{equation}
where \(\tau \in \mathbb{R}\) is a temperature parameter. We define the critic function \(\theta(\cdot, \cdot)\) by
\[
	\theta(\bm{h}_i, \bm{h}_j) = \frac{g(\bm{h}_i)^\top g(\bm{h}_j)}{\|g(\bm{h}_i)\| \|g(\bm{h}_i)\|},
\]
where \(g(\cdot)\) is parameterized by a non-linear multilayer perceptron to enhance expressive power \cite{Chen:2020wj}.

\subsection{Structure-Aware Hard Negative Mining}
\label{sec:hard-negative-mining}

\begin{figure}
	\centering
	\includegraphics[width=0.8\linewidth]{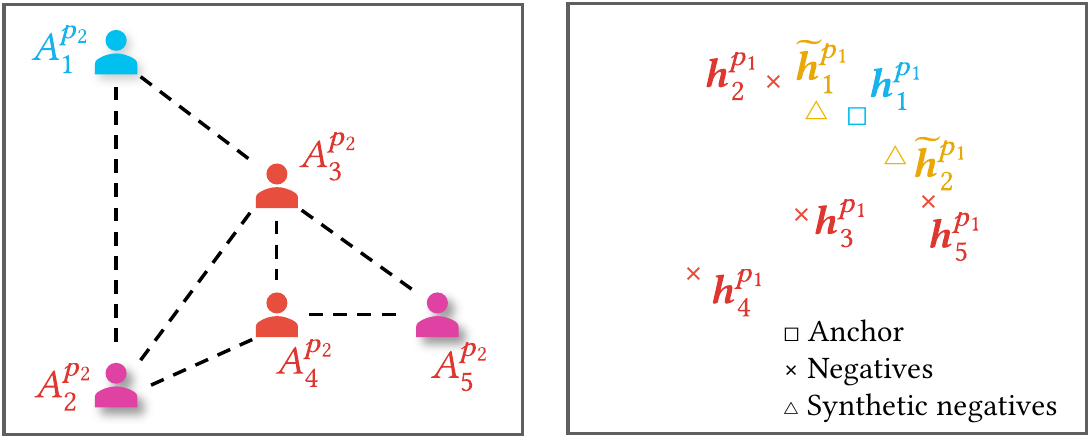}
	\caption{The proposed structure-aware hard negative mining scheme, which discovers structurally hard negatives in each view and further synthesizes additional harder negative samples.}
	\label{fig:hard-negative-mining}
\end{figure}

As with previous studies \cite{Schroff:2015wo,Cai:2020tz,Xuan:2020is}, CL benefits from hard negative samples, i.e. samples close to the anchor node such that cannot be distinguished easily.
In the context of HGs, we observe that semantic-level node representations are not sufficient to calculate the hardness of each negative pair. Therefore, in this work, to effectively measure hardness of each sample with respect to the anchor, we propose to explore the hardness of negative samples in terms of their structural similarities.
The proposed structure-aware hard negative mining scheme is illustrated in Figure \ref{fig:hard-negative-mining}.

We first introduce a structure-aware metric \(s(i, j, p)\) representing distance measure of a negative node \(v_i\) to the anchor node \(v_j\) given a semantic view \(p\), which can be regarded as the hardness of the negative node \(v_i\).
Note that in order to empower the model with inductive capabilities, we prefer a local measure to a global one.
In this paper, we propose two model variants \themodel-PPR and \themodel-PE, which use Laplacian positional embeddings and personalized PageRank scores for structure-aware hard negative mining respectively.
\begin{itemize}
	\item The Personalized PageRank (PPR) score \cite{Jeh:2003kz,Page:1999wg} of node \(v\) is defined as the stationary distribution of a random walk starting from and returning to node \(v\) at a probability of \(c\) at each step. Formally, the PPR vector of node \(v\) under semantic view \(p\) satisfies the following equation
		\begin{equation}
			\bm{s}_v^p = (1 - c) \bm{A}^p \bm{s}_v^p + c \bm{I} \bm{p}_v,
		\end{equation}
	where \(c\) is the returning probability and \(\bm p_v\) is the preference vector with \((\bm{p}_v)_i = 1\) when \(i = v\) and all other entries set to \(0\). \(\bm{A}^p\) denotes the adjacency matrix generated by metapath \(p\). The structural similarity between node \(v\) and \(k\) can be represented by the PPR score of node \(k\) with respect to node \(v\), i.e. \((\bm{s}^p_v)_k\).
	\item The Laplacian positional embedding of one node is defined to be its \(k\) smallest non-trivial eigenvectors \cite{Dwivedi:2020ws}. We simply define the structure similarity as the inner product between \(\bm{s}_i^p\) and \(\bm{s}_j^p\).
\end{itemize}

After that, we perform hard negative mining by giving larger weights to harder negative samples. Specifically, we sort negatives according to the hardness metric and pick the top-\(T\) negatives to form a candidate list for semantic view \(p\). Then, we synthesize \(M \ll |\mathcal{V}|\) samples by creating a convex linear combination of them. The generated sample \(\widetilde{\bm{h}}_m^p\) is mathematically expressed as
\begin{equation}
	\widetilde{\bm{h}}_m^p = \alpha_m \bm{h}_i^p + (1 - \alpha_m) \bm{h}_j^p,
	\label{eq:mixup}
\end{equation}
where \(\bm{h}_i^p, \bm{h}_j^p \in \mathcal{B}^p\) are randomly picked from the memory bank, \(\alpha_m \sim \mathrm{Beta}(\alpha, \alpha)\), and \(\alpha\) is a hyperparameter, fixed to \(1\) in our experiments.
These interpolated samples will be added into negative bank when estimating mutual information \(I(\bm{h}_i^p; \bm{h}_i)\), as given in sequel
\begin{equation}
	\mathcal{L}(\bm{h}^p_i, \bm{h}_i) = - \log\frac{ e^{\theta(\bm{h}^p_{i}, \bm{h}_{i}) / \tau}}
			{e^{\theta(\bm{h}^p_{i}, \bm{h}_{i}) / \tau} + \sum\limits_{\bm{h} \in \mathcal{B}^p} e^{\theta(\bm{h}^p_{i}, \bm{h}) / \tau} },
	\label{eq:final-contrastive-objective}
\end{equation}
where the negative bank 
\begin{equation}
	\mathcal{B}^p = \{\bm{h}^p_j\}_{j \neq i} \cup \{\bm{h}_j\}_{j \neq i} \cup \{\widetilde{\bm{h}}_m^p\}_{m=1}^M
	\label{eq:negative-bank}
\end{equation}
consists of all inter-view and intra-view negatives as well as synthesized hard negatives.
The contrastive objective \(\ell(\bm{h}_i; \bm{h}^p_i)\) for the aggregated node representation \(\bm{h}_i\) can be defined similarly as Eq. (\ref{eq:final-contrastive-objective}).
The final objective is an average of the losses from all contrastive pairs, formally given by
\begin{equation}
  \mathcal{J} = \frac{1}{|\mathcal{V}|} \sum_{v_i \in \mathcal{V}} \left[ \frac{1}{|\mathcal{P}|} \sum_{p \in \mathcal{P}} \frac{1}{2} \left( \mathcal{L}(\bm{h}^p_i; \bm{h}_i) + \mathcal{L}(\bm{h}_i; \bm{h}_i^p) \right) \right].
  \label{eq:objective}
\end{equation}

\subsection{Complexity Analysis}

Most computational burden of the \themodel framework lies in the contrastive objective, which involves computing \((|\mathcal{V}|^2 |\mathcal{P}|)\) node embedding pairs. For structure-aware hard negative mining, the synthesized samples incur an additional computational cost of \(O(M|\mathcal{V}||\mathcal{P}|)\), which is equivalent to increasing the memory size by \(M \ll |\mathcal{V}|\).
The construction of the candidates list of hard negatives only depends on graph structures of each semantic view, and thus it can be regarded as a preprocessing process.

\begin{table}
	\caption{Performance comparison on three datasets. Node classification performance is in terms of Macro-F1 (Ma-F1) and Micro-F1 (Mi-F1). The highest performance of unsupervised and supervised models is boldfaced and underlined, respectively.}
	\centering
	\resizebox{\linewidth}{!}{
	\begin{tabular}{ccccccc}
	\toprule
	\multirow{2.5}{*}{Method} & \multicolumn{2}{c}{ACM} & \multicolumn{2}{c}{IMDb} & \multicolumn{2}{c}{DBLP} \\
	\cmidrule(lr){2-3} \cmidrule(lr){4-5} \cmidrule(lr){6-7}
          &    Mi-F1 & Ma-F1 & Mi-F1 & Ma-F1 & Mi-F1 & Ma-F1 \\
    \midrule
    DeepWalk & 76.92  & 77.25  & 46.38  & 40.72  & 79.37  & 77.43 \\
    ESim  & 76.89  & 77.32  & 35.28  & 32.10  & 92.73  & 91.64 \\
    metapath2vec   & 65.00  & 65.09  & 45.65  & 41.16  & 91.53  & 90.76 \\
    HERec & 66.03  & 66.17  & 45.81  & 41.65  & 92.69  & 91.78 \\
	\midrule
    HAN-U & 82.63  & 81.89  & 43.98  & 40.87  & 90.47  & 89.65 \\
    DGI   & 89.15  & 89.09  & 48.86  & 45.38  & 91.30  & 90.69 \\
    GRACE  & 88.72  & 88.72  & 46.64  & 42.41  & 90.88  & 89.76 \\
    \rowcolor{gray!20}\themodel-PE & \textbf{90.76} & \textbf{90.72} & \textbf{58.98} & \textbf{54.48} & \textbf{92.81} & \textbf{92.33} \\
	\rowcolor{gray!20}\themodel-PPR & 90.75  & 90.70  & 58.96  & 54.47  & 92.78  & 92.30 \\ 
    \specialrule{0.5pt}{0.5pt}{1pt}
	\midrule
    GCN   & 86.77  & 86.81  & 49.78  & 45.73  & 91.71  & 90.79 \\
    GAT   & 86.01  & 86.23  & 55.28  & 49.44  & 91.96  & 90.97 \\
    HAN   & \underline{89.22}  & \underline{89.40}  & \underline{54.17}  & \underline{49.78}  & \underline{92.05}  & \underline{91.17} \\
    \bottomrule
    \end{tabular}
    }
	\label{tab:performance}
\end{table}

\section{Experiments}

We evaluate the effectiveness of our proposed \themodel in this section.
The purpose of empirical studies is to answer the following questions.
\begin{itemize}
	\item \textbf{RQ1}. How does our proposed \themodel outperform other representative baseline algorithms?
	\item \textbf{RQ2}. How does the proposed structure-aware hard negative mining scheme affect the performance of \themodel?
\end{itemize}

\subsection{Experimental Configurations}

\subsubsection{Datasets}
To achieve a comprehensive comparison, we use three widely-used heterogeneous datasets from different domains: DBLP, ACM, and IMDb, where DBLP and ACM are two academic networks, and IMDb is a movie network.
\begin{itemize}
	\item \textbf{DBLP} is a subset of an academic network extracted from DBLP, consisting of four kinds of nodes: authors, papers, conferences, and topics. Each author is labeled with their research area.
	\item \textbf{ACM} is an academic network from ACM. We construct a heterogeneous graph with nodes of three types: papers, authors, and subjects. Papers are labeled by their research topic. 
	\item \textbf{IMDb} is a subset of the movie network IMDb, where nodes represent movies, actors, or directors. We categorize movies into three classes according to their genre. 
\end{itemize}

\subsubsection{Baselines}
We compare the proposed \themodel against a comprehensive set of baselines, including both representative traditional and deep graph representation learning methods.
These baselines include (1) DeepWalk \cite{Perozzi:2014ib}, DGI \cite{Velickovic:2018we}, GRACE \cite{Zhu:2020vf}, GCN \cite{Kipf:2017tc}, and GAT \cite{Velickovic:2019tu}, designed for homogeneous graphs, and (2) ESim \cite{Shang:2016wf}, HERec \cite{Shi:2019kf}, and HAN \cite{Wang:2019gv} for heterogeneous graphs.
Among them, DeepWalk, ESim, metapath2vec, and HERec only utilize metapaths for generating multiple views for training. HAN and DGI further use associated node features (if any) for learning representations. Furthermore, we also include three representative \emph{supervised} baselines GCN, GAT, and HAN for reference.
Following HAN \cite{Wang:2019gv}, for DeepWalk, we simply discard node and edge types, and treat the heterogeneous graph as a homogeneous graph; for DGI, GRACE, GCN, and GAT, we generate homogeneous graphs according to all metapaths, and report the best performance.

\subsubsection{Evaluation protocols}
For comprehensive evaluation, we follow HAN \cite{Wang:2019gv} and perform experiments on the node classification task. 
For node classification, we run a \(k\)-NN classifier with \(k=5\) on the learned node embeddings. We report performance in terms of Micro-F1 and Macro-F1 for evaluation of node classification. For dataset split, we randomly pick 20\% nodes in each dataset for training and the remaining 80\% for test. Results from 10 different random splits are averaged for the final report.

\subsection{Performance Comparison (RQ1)}

Experiment results are presented in Table \ref{tab:performance}. Overall, our proposed \themodel achieves the best unsupervised performance on almost all datasets. It is worth mentioning that our \themodel is competitive to and even better than several supervised counterparts.

Compared with traditional approaches based on random walks and matrix decomposition, our proposed GNN-based \themodel outperforms them by large margins. Particularly, \themodel improves metapath2vec and HERec by over 25\% on ACM, which demonstrates the superiority of GNN that can leverage rich node attributes to learn high quality node representations for heterogeneous graphs. 

For deep unsupervised learning methods, our \themodel achieves promising improvements as well. For the unsupervised version HAN-U that is trained with a simple reconstruction loss, its performance is even inferior to HERec on IMDb and DBLP despite its utilization of node attributes. This indicates that the reconstruction loss is insufficient to fully exploit the structural and semantic information for node-centric tasks such as node classification and clustering.
Compared to DGI, a homogeneous contrastive learning method, \themodel accomplishes excelled performance on all datasets and evaluation tasks, especially on ACM and IMDb dataset, where large improvements on both tasks are achieved. This validates the effectiveness of our proposed view-to-aggregation contrastive objective and structure-aware hard negative mining strategy.

Furthermore, experiments show that \themodel even outperforms its supervised baselines on ACM and IMDb datasets. It remarkably improves HAN by over 4\% in terms of node classification Micro-F1 score on IMDb. This outstanding performance of \themodel certifies the superiority of our proposed heterogeneous graph contrastive learning framework such that it can distill useful information from each semantic view.

\subsection{Close Inspections on Structure-Aware Hard Negative Mining Module (RQ2)}

\subsubsection{Effectiveness of the module}
We modify the negative bank in our contrastive objective to study the impact of structure-aware hard negative mining component. \themodel-- denotes the model with synthesized harder samples \(\{\widetilde{\bm{h}}_m^p\}_{m=1}^M\) removed, where the negative bank 
\(
	\mathcal{B}^p = \{\bm{h}^p_j\}_{j \neq i} \cup \{\bm{h}_j\}_{j \neq i} 
	\label{eq:negative-bank}
\)
consists of only inter-view and intra-view negatives.
We also construct a model variant \themodel-Sem, that discovers and synthesizes semantic negative samples using inner product of node embeddings.

The results are presented in Table \ref{tab:ablation}. It is observed that \themodel improves consistently than \themodel-- on all datasets for both node classification and clustering tasks. Especially for node clustering task on ACM, the gain reaches up to 15\%. This verifies the effectiveness of our synthesizing hard negative sample strategy: giving larger weights to harder negative samples with the delicately designed synthesis term. Furthermore, the outstanding performance of \themodel compared to the model variant \themodel-Sem further justifies the superiority of structural-aware hard negative mining.

\subsubsection{Hyperparameters sensitivity analysis.}
\begin{figure}
	\centering
	\subfloat[Synthesized hard negatives]{
		\includegraphics[width=0.48\linewidth]{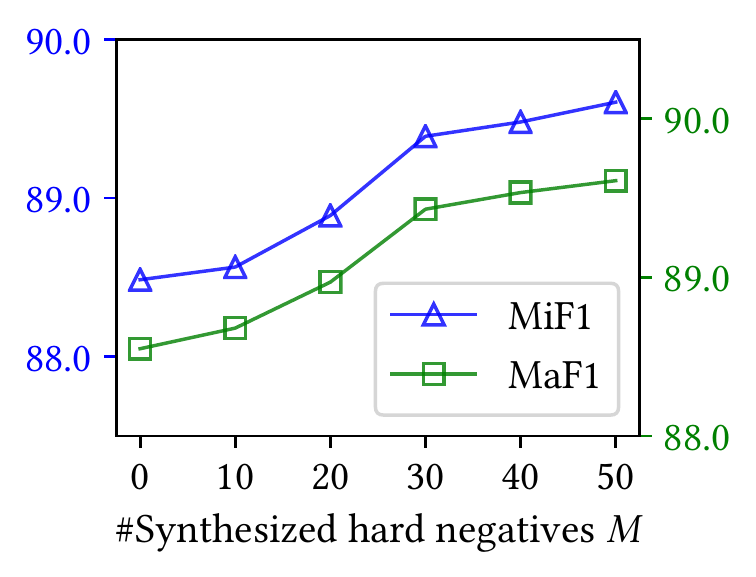}
		\label{fig:sensitivity-mixup}
	}\hfill
	\subfloat[Candidate hard negative samples]{
		\includegraphics[width=0.48\linewidth]{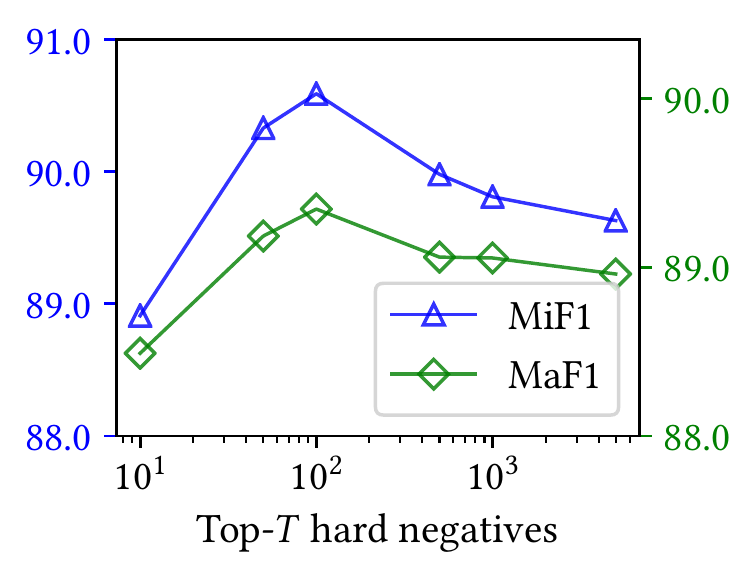}
		\label{fig:sensitivity-topt-clf}
	}
	\caption{Model performance with varied numbers of synthesized hard negatives and candidate hard negative samples \(T\) on the ACM dataset.}
\end{figure}

We study how the two key parameters in the hard negative mining module affect the performance of HORACE: the number of synthesized hard negatives \(M\) and the threshold \(T\) in selecting top-\(T\) candidate hard negatives.
Results on ACM under different parameter settings are obtained and reported by only varying one specific parameter and keeping all other parameters the same.
As is shown in Figure \ref{fig:sensitivity-mixup}, 
the performance level of HORACE increases as the number of synthesized negatives \(M\) increases. This indicates that the learning of HORACE benefits from the synthesized hard negatives.
For the parameter \(T\), as presented in Figure \ref{fig:sensitivity-topt-clf}, the model performance first rises with a larger \(T\), but soon the performance levels off and decreases as \(T\) increases further. We suspect that this is because a larger \(T\) will result in the selection of less hard negatives, reducing the benefits brought by our proposed hard negative sampling strategy.

\begin{table}[t]
	\centering
	\caption{Effectiveness of the structure-aware hard negative mining module.}
	\resizebox{\linewidth}{!}{
    \begin{tabular}{ccccccc}
    \toprule
	\multirow{2.5}{*}{Method} & \multicolumn{2}{c}{ACM} & \multicolumn{2}{c}{IMDb} & \multicolumn{2}{c}{DBLP} \\
	\cmidrule(lr){2-3} \cmidrule(lr){4-5} \cmidrule(lr){6-7}
	          & Mi-F1 & Ma-F1 & Mi-F1 & Ma-F1 & Mi-F1 & Ma-F1 \\
    \midrule
    \themodel-- & 88.62  & 88.43  & 57.94  & 52.97  & 92.42  & 91.85 \\
    \themodel-Sem & 90.24  & 90.18  & 58.95  & 52.38  & 92.73  & 92.21 \\
    \rowcolor{gray!20}\themodel-PE & \textbf{91.40} & \textbf{91.45} & \textbf{58.96} & \textbf{53.73} & \textbf{92.77} & \textbf{92.28} \\
    \bottomrule
    \end{tabular}
    }
  \label{tab:ablation}
\end{table}

\section{Conclusion}
This paper has developed a novel heterogeneous graph contrastive learning framework.
To alleviate the label scarcity problem, we leverage contrastive learning techniques that enables self-supervised training for HGs. 
We further propose a novel hard negative mining scheme to improve the embedding quality, considering the complex structure of heterogeneous graphs and smoothing nature of heterogeneous GNNs. The proposed structure-aware negative mining scheme discovers and reweights structurally hard negatives so that they contribute more to contrastive learning.
Extensive experiments have been conducted on three real-world heterogeneous datasets. The experimental results show that our proposed method not only consistently outperforms representative unsupervised baseline methods, but also achieves on par performance with supervised counterparts, and is even superior to several of them.

\bibliographystyle{ACM-Reference-Format}
\bibliography{kdd2021}


\begin{thebibliography}{27}


\ifx \showCODEN    \undefined \def \showCODEN     #1{\unskip}     \fi
\ifx \showDOI      \undefined \def \showDOI       #1{#1}\fi
\ifx \showISBNx    \undefined \def \showISBNx     #1{\unskip}     \fi
\ifx \showISBNxiii \undefined \def \showISBNxiii  #1{\unskip}     \fi
\ifx \showISSN     \undefined \def \showISSN      #1{\unskip}     \fi
\ifx \showLCCN     \undefined \def \showLCCN      #1{\unskip}     \fi
\ifx \shownote     \undefined \def \shownote      #1{#1}          \fi
\ifx \showarticletitle \undefined \def \showarticletitle #1{#1}   \fi
\ifx \showURL      \undefined \def \showURL       {\relax}        \fi
\providecommand\bibfield[2]{#2}
\providecommand\bibinfo[2]{#2}
\providecommand\natexlab[1]{#1}
\providecommand\showeprint[2][]{arXiv:#2}

\bibitem[\protect\citeauthoryear{Cai, Frankle, Schwab, and Morcos}{Cai
  et~al\mbox{.}}{2020}]%
        {Cai:2020tz}
\bibfield{author}{\bibinfo{person}{Tiffany~Tianhui Cai},
  \bibinfo{person}{Jonathan Frankle}, \bibinfo{person}{David~J Schwab}, {and}
  \bibinfo{person}{Ari~S Morcos}.} \bibinfo{year}{2020}\natexlab{}.
\newblock \showarticletitle{{Are All Negatives Created Equal in Contrastive
  Instance Discrimination?}}
\newblock \bibinfo{journal}{\emph{arXiv.org}} (\bibinfo{date}{Oct.}
  \bibinfo{year}{2020}).
\newblock
\showeprint[arxiv]{2010.06682v2}~[cs.CV]


\bibitem[\protect\citeauthoryear{Caron, Misra, Mairal, Goyal, Bojanowski, and
  Joulin}{Caron et~al\mbox{.}}{2020}]%
        {Caron:2020uv}
\bibfield{author}{\bibinfo{person}{Mathilde Caron}, \bibinfo{person}{Ishan
  Misra}, \bibinfo{person}{Julien Mairal}, \bibinfo{person}{Priya Goyal},
  \bibinfo{person}{Piotr Bojanowski}, {and} \bibinfo{person}{Armand Joulin}.}
  \bibinfo{year}{2020}\natexlab{}.
\newblock \showarticletitle{{Unsupervised Learning of Visual Features by
  Contrasting Cluster Assignments}}. In \bibinfo{booktitle}{\emph{NeurIPS}}.
\newblock


\bibitem[\protect\citeauthoryear{Chen, Kornblith, Norouzi, and Hinton}{Chen
  et~al\mbox{.}}{2020}]%
        {Chen:2020wj}
\bibfield{author}{\bibinfo{person}{Ting Chen}, \bibinfo{person}{Simon
  Kornblith}, \bibinfo{person}{Mohammad Norouzi}, {and}
  \bibinfo{person}{Geoffrey~E. Hinton}.} \bibinfo{year}{2020}\natexlab{}.
\newblock \showarticletitle{{A Simple Framework for Contrastive Learning of
  Visual Representations}}. In \bibinfo{booktitle}{\emph{ICML}}.
  \bibinfo{pages}{1597--1607}.
\newblock


\bibitem[\protect\citeauthoryear{Chen and He}{Chen and He}{2020}]%
        {Chen:2020uu}
\bibfield{author}{\bibinfo{person}{Xinlei Chen} {and} \bibinfo{person}{Kaiming
  He}.} \bibinfo{year}{2020}\natexlab{}.
\newblock \showarticletitle{{Exploring Simple Siamese Representation
  Learning}}.
\newblock \bibinfo{journal}{\emph{arXiv.org}} (\bibinfo{date}{Nov.}
  \bibinfo{year}{2020}).
\newblock
\showeprint[arxiv]{2011.10566v1}~[cs.CV]


\bibitem[\protect\citeauthoryear{Dwivedi, Joshi, Laurent, Bengio, and
  Bresson}{Dwivedi et~al\mbox{.}}{2020}]%
        {Dwivedi:2020ws}
\bibfield{author}{\bibinfo{person}{Vijay~Prakash Dwivedi},
  \bibinfo{person}{Chaitanya~K. Joshi}, \bibinfo{person}{Thomas Laurent},
  \bibinfo{person}{Yoshua Bengio}, {and} \bibinfo{person}{Xavier Bresson}.}
  \bibinfo{year}{2020}\natexlab{}.
\newblock \showarticletitle{{Benchmarking Graph Neural Networks}}.
\newblock \bibinfo{journal}{\emph{arXiv.org}} (\bibinfo{date}{March}
  \bibinfo{year}{2020}).
\newblock
\showeprint[arxiv]{2003.00982v3}~[cs.LG]


\bibitem[\protect\citeauthoryear{Fu, Zhang, Meng, and King}{Fu
  et~al\mbox{.}}{2020}]%
        {Fu:2020fs}
\bibfield{author}{\bibinfo{person}{Xinyu Fu}, \bibinfo{person}{Jiani Zhang},
  \bibinfo{person}{Ziqiao Meng}, {and} \bibinfo{person}{Irwin King}.}
  \bibinfo{year}{2020}\natexlab{}.
\newblock \showarticletitle{{MAGNN: Metapath Aggregated Graph Neural Network
  for Heterogeneous Graph Embedding}}. In \bibinfo{booktitle}{\emph{WWW}}.
  \bibinfo{pages}{2331--2341}.
\newblock


\bibitem[\protect\citeauthoryear{Grill, Strub, Altch{\'e}, Tallec, Richemond,
  Buchatskaya, Doersch, Pires, Guo, Azar, Piot, Kavukcuoglu, Munos, and
  Valko}{Grill et~al\mbox{.}}{2020}]%
        {Grill:2020uc}
\bibfield{author}{\bibinfo{person}{Jean-Bastien Grill},
  \bibinfo{person}{Florian Strub}, \bibinfo{person}{Florent Altch{\'e}},
  \bibinfo{person}{Corentin Tallec}, \bibinfo{person}{Pierre~H. Richemond},
  \bibinfo{person}{Elena Buchatskaya}, \bibinfo{person}{Carl Doersch},
  \bibinfo{person}{Bernardo~Avila Pires}, \bibinfo{person}{Zhaohan~Daniel Guo},
  \bibinfo{person}{Mohammad~Gheshlaghi Azar}, \bibinfo{person}{Bilal Piot},
  \bibinfo{person}{Koray Kavukcuoglu}, \bibinfo{person}{R{\'e}mi Munos}, {and}
  \bibinfo{person}{Michal Valko}.} \bibinfo{year}{2020}\natexlab{}.
\newblock \showarticletitle{{Bootstrap Your Own Latent: A New Approach to
  Self-Supervised Learning}}. In \bibinfo{booktitle}{\emph{NeurIPS}}.
\newblock


\bibitem[\protect\citeauthoryear{Hassani and Khasahmadi}{Hassani and
  Khasahmadi}{2020}]%
        {Hassani:2020un}
\bibfield{author}{\bibinfo{person}{Kaveh Hassani} {and}
  \bibinfo{person}{Amir~Hosein Khasahmadi}.} \bibinfo{year}{2020}\natexlab{}.
\newblock \showarticletitle{{Contrastive Multi-View Representation Learning on
  Graphs}}. In \bibinfo{booktitle}{\emph{ICML}}. \bibinfo{pages}{4116--4126}.
\newblock


\bibitem[\protect\citeauthoryear{H{\'e}naff, Srinivas, De~Fauw, Razavi,
  Doersch, Eslami, and van~den Oord}{H{\'e}naff et~al\mbox{.}}{2020}]%
        {Henaff:2020ta}
\bibfield{author}{\bibinfo{person}{Olivier~J. H{\'e}naff},
  \bibinfo{person}{Aravind Srinivas}, \bibinfo{person}{Jeffrey De~Fauw},
  \bibinfo{person}{Ali Razavi}, \bibinfo{person}{Carl Doersch},
  \bibinfo{person}{S.~M.~Ali Eslami}, {and} \bibinfo{person}{A{\"a}ron van~den
  Oord}.} \bibinfo{year}{2020}\natexlab{}.
\newblock \showarticletitle{{Data-Efficient Image Recognition with Contrastive
  Predictive Coding}}. In \bibinfo{booktitle}{\emph{ICML}}.
  \bibinfo{pages}{4182--4192}.
\newblock


\bibitem[\protect\citeauthoryear{Jeh and Widom}{Jeh and Widom}{2003}]%
        {Jeh:2003kz}
\bibfield{author}{\bibinfo{person}{Glen Jeh} {and} \bibinfo{person}{Jennifer
  Widom}.} \bibinfo{year}{2003}\natexlab{}.
\newblock \showarticletitle{{Scaling Personalized Web Search}}. In
  \bibinfo{booktitle}{\emph{WWW}}.
\newblock


\bibitem[\protect\citeauthoryear{Kipf and Welling}{Kipf and Welling}{2017}]%
        {Kipf:2017tc}
\bibfield{author}{\bibinfo{person}{Thomas~N. Kipf} {and} \bibinfo{person}{Max
  Welling}.} \bibinfo{year}{2017}\natexlab{}.
\newblock \showarticletitle{{Semi-Supervised Classification with Graph
  Convolutional Networks}}. In \bibinfo{booktitle}{\emph{ICLR}}.
\newblock


\bibitem[\protect\citeauthoryear{Page, Brin, Motwani, and Winograd}{Page
  et~al\mbox{.}}{1999}]%
        {Page:1999wg}
\bibfield{author}{\bibinfo{person}{Lawrence Page}, \bibinfo{person}{Sergey
  Brin}, \bibinfo{person}{Rajeev Motwani}, {and} \bibinfo{person}{Terry
  Winograd}.} \bibinfo{year}{1999}\natexlab{}.
\newblock \bibinfo{booktitle}{\emph{{The PageRank Citation Ranking: Bringing
  Order to the Web}}}.
\newblock \bibinfo{type}{{T}echnical {R}eport}.
\newblock


\bibitem[\protect\citeauthoryear{Perozzi, Al-Rfou, and Skiena}{Perozzi
  et~al\mbox{.}}{2014}]%
        {Perozzi:2014ib}
\bibfield{author}{\bibinfo{person}{Bryan Perozzi}, \bibinfo{person}{Rami
  Al-Rfou}, {and} \bibinfo{person}{Steven Skiena}.}
  \bibinfo{year}{2014}\natexlab{}.
\newblock \showarticletitle{{DeepWalk: Online Learning of Social
  Representations}}. In \bibinfo{booktitle}{\emph{KDD}}.
  \bibinfo{pages}{701--710}.
\newblock


\bibitem[\protect\citeauthoryear{Schroff, Kalenichenko, and Philbin}{Schroff
  et~al\mbox{.}}{2015}]%
        {Schroff:2015wo}
\bibfield{author}{\bibinfo{person}{Florian Schroff}, \bibinfo{person}{Dmitry
  Kalenichenko}, {and} \bibinfo{person}{James Philbin}.}
  \bibinfo{year}{2015}\natexlab{}.
\newblock \showarticletitle{{FaceNet: A Unified Embedding for Face Recognition
  and Clustering}}. In \bibinfo{booktitle}{\emph{CVPR}}.
  \bibinfo{pages}{815--823}.
\newblock


\bibitem[\protect\citeauthoryear{Shang, Qu, Liu, Kaplan, Han, and Peng}{Shang
  et~al\mbox{.}}{2016}]%
        {Shang:2016wf}
\bibfield{author}{\bibinfo{person}{Jingbo Shang}, \bibinfo{person}{Meng Qu},
  \bibinfo{person}{Jialu Liu}, \bibinfo{person}{Lance~M Kaplan},
  \bibinfo{person}{Jiawei Han}, {and} \bibinfo{person}{Jian Peng}.}
  \bibinfo{year}{2016}\natexlab{}.
\newblock \showarticletitle{{Meta-Path Guided Embedding for Similarity Search
  in Large-Scale Heterogeneous Information Networks}}.
\newblock \bibinfo{journal}{\emph{arXiv.org}} (\bibinfo{date}{Oct.}
  \bibinfo{year}{2016}).
\newblock
\showeprint[arxiv]{1610.09769v1}~[cs.SI]


\bibitem[\protect\citeauthoryear{Shi, Hu, Zhao, and Yu}{Shi
  et~al\mbox{.}}{2019}]%
        {Shi:2019kf}
\bibfield{author}{\bibinfo{person}{Chuan Shi}, \bibinfo{person}{Binbin Hu},
  \bibinfo{person}{Wayne~Xin Zhao}, {and} \bibinfo{person}{Philip~S. Yu}.}
  \bibinfo{year}{2019}\natexlab{}.
\newblock \showarticletitle{{Heterogeneous Information Network Embedding for
  Recommendation}}.
\newblock \bibinfo{journal}{\emph{TKDE}} \bibinfo{volume}{31},
  \bibinfo{number}{2} (\bibinfo{year}{2019}), \bibinfo{pages}{357--370}.
\newblock


\bibitem[\protect\citeauthoryear{Sun, Barber, Gupta, Aggarwal, and Han}{Sun
  et~al\mbox{.}}{2011}]%
        {Sun:2011be}
\bibfield{author}{\bibinfo{person}{Yizhou Sun}, \bibinfo{person}{Rick Barber},
  \bibinfo{person}{Manish Gupta}, \bibinfo{person}{Charu~C. Aggarwal}, {and}
  \bibinfo{person}{Jiawei Han}.} \bibinfo{year}{2011}\natexlab{}.
\newblock \showarticletitle{{Co-author Relationship Prediction in Heterogeneous
  Bibliographic Networks}}. In \bibinfo{booktitle}{\emph{ASONAM}}.
  \bibinfo{pages}{121--128}.
\newblock


\bibitem[\protect\citeauthoryear{Tian, Krishnan, and Isola}{Tian
  et~al\mbox{.}}{2020a}]%
        {Tian:2020tz}
\bibfield{author}{\bibinfo{person}{Yonglong Tian}, \bibinfo{person}{Dilip
  Krishnan}, {and} \bibinfo{person}{Phillip Isola}.}
  \bibinfo{year}{2020}\natexlab{a}.
\newblock \showarticletitle{{Contrastive Representation Distillation}}. In
  \bibinfo{booktitle}{\emph{ICLR}}.
\newblock


\bibitem[\protect\citeauthoryear{Tian, Sun, Poole, Krishnan, Schmid, and
  Isola}{Tian et~al\mbox{.}}{2020b}]%
        {Tian:2020vw}
\bibfield{author}{\bibinfo{person}{Yonglong Tian}, \bibinfo{person}{Chen Sun},
  \bibinfo{person}{Ben Poole}, \bibinfo{person}{Dilip Krishnan},
  \bibinfo{person}{Cordelia Schmid}, {and} \bibinfo{person}{Phillip Isola}.}
  \bibinfo{year}{2020}\natexlab{b}.
\newblock \showarticletitle{{What Makes for Good Views for Contrastive
  Learning}}. In \bibinfo{booktitle}{\emph{NeurIPS}}.
\newblock


\bibitem[\protect\citeauthoryear{van~den Oord, Li, and Vinyals}{van~den Oord
  et~al\mbox{.}}{2018}]%
        {vandenOord:2018ut}
\bibfield{author}{\bibinfo{person}{A{\"a}ron van~den Oord},
  \bibinfo{person}{Yazhe Li}, {and} \bibinfo{person}{Oriol Vinyals}.}
  \bibinfo{year}{2018}\natexlab{}.
\newblock \showarticletitle{{Representation Learning with Contrastive
  Predictive Coding}}.
\newblock \bibinfo{journal}{\emph{arXiv.org}} (\bibinfo{year}{2018}).
\newblock
\showeprint[arxiv]{1807.03748v2}~[cs.LG]


\bibitem[\protect\citeauthoryear{Veli{\v c}kovi{\'c}, Cucurull, Casanova,
  Romero, Li{\`o}, and Bengio}{Veli{\v c}kovi{\'c} et~al\mbox{.}}{2018}]%
        {Velickovic:2018we}
\bibfield{author}{\bibinfo{person}{Petar Veli{\v c}kovi{\'c}},
  \bibinfo{person}{Guillem Cucurull}, \bibinfo{person}{Arantxa Casanova},
  \bibinfo{person}{Adriana Romero}, \bibinfo{person}{Pietro Li{\`o}}, {and}
  \bibinfo{person}{Yoshua Bengio}.} \bibinfo{year}{2018}\natexlab{}.
\newblock \showarticletitle{{Graph Attention Networks}}. In
  \bibinfo{booktitle}{\emph{ICLR}}.
\newblock


\bibitem[\protect\citeauthoryear{Veli{\v c}kovi{\'c}, Fedus, Hamilton, Li{\`o},
  Bengio, and Hjelm}{Veli{\v c}kovi{\'c} et~al\mbox{.}}{2019}]%
        {Velickovic:2019tu}
\bibfield{author}{\bibinfo{person}{Petar Veli{\v c}kovi{\'c}},
  \bibinfo{person}{William Fedus}, \bibinfo{person}{William~L. Hamilton},
  \bibinfo{person}{Pietro Li{\`o}}, \bibinfo{person}{Yoshua Bengio}, {and}
  \bibinfo{person}{R.~Devon Hjelm}.} \bibinfo{year}{2019}\natexlab{}.
\newblock \showarticletitle{{Deep Graph Infomax}}. In
  \bibinfo{booktitle}{\emph{ICLR}}.
\newblock


\bibitem[\protect\citeauthoryear{Wang, Ji, Shi, Wang, Ye, Cui, and Yu}{Wang
  et~al\mbox{.}}{2019}]%
        {Wang:2019gv}
\bibfield{author}{\bibinfo{person}{Xiao Wang}, \bibinfo{person}{Houye Ji},
  \bibinfo{person}{Chuan Shi}, \bibinfo{person}{Bai Wang},
  \bibinfo{person}{Yanfang Ye}, \bibinfo{person}{Peng Cui}, {and}
  \bibinfo{person}{Philip~S. Yu}.} \bibinfo{year}{2019}\natexlab{}.
\newblock \showarticletitle{{Heterogeneous Graph Attention Network}}. In
  \bibinfo{booktitle}{\emph{WWW}}. \bibinfo{pages}{2022--2032}.
\newblock


\bibitem[\protect\citeauthoryear{Xuan, Stylianou, Liu, and Pless}{Xuan
  et~al\mbox{.}}{2020}]%
        {Xuan:2020is}
\bibfield{author}{\bibinfo{person}{Hong Xuan}, \bibinfo{person}{Abby
  Stylianou}, \bibinfo{person}{Xiaotong Liu}, {and} \bibinfo{person}{Robert
  Pless}.} \bibinfo{year}{2020}\natexlab{}.
\newblock \showarticletitle{{Hard Negative Examples are Hard, but Useful}}. In
  \bibinfo{booktitle}{\emph{ECCV}}. \bibinfo{pages}{126--142}.
\newblock


\bibitem[\protect\citeauthoryear{You, Chen, Sui, Chen, Wang, and Shen}{You
  et~al\mbox{.}}{2020}]%
        {You:2020ut}
\bibfield{author}{\bibinfo{person}{Yuning You}, \bibinfo{person}{Tianlong
  Chen}, \bibinfo{person}{Yongduo Sui}, \bibinfo{person}{Ting Chen},
  \bibinfo{person}{Zhangyang Wang}, {and} \bibinfo{person}{Yang Shen}.}
  \bibinfo{year}{2020}\natexlab{}.
\newblock \showarticletitle{{Graph Contrastive Learning with Augmentations}}.
  In \bibinfo{booktitle}{\emph{NeurIPS}}.
\newblock


\bibitem[\protect\citeauthoryear{Zhu, Xu, Yu, Liu, Wu, and Wang}{Zhu
  et~al\mbox{.}}{2020}]%
        {Zhu:2020vf}
\bibfield{author}{\bibinfo{person}{Yanqiao Zhu}, \bibinfo{person}{Yichen Xu},
  \bibinfo{person}{Feng Yu}, \bibinfo{person}{Qiang Liu}, \bibinfo{person}{Shu
  Wu}, {and} \bibinfo{person}{Liang Wang}.} \bibinfo{year}{2020}\natexlab{}.
\newblock \showarticletitle{{Deep Graph Contrastive Representation Learning}}.
  In \bibinfo{booktitle}{\emph{GRL+@ICML}}.
\newblock


\bibitem[\protect\citeauthoryear{Zhu, Xu, Yu, Liu, Wu, and Wang}{Zhu
  et~al\mbox{.}}{2021}]%
        {Zhu:2021wh}
\bibfield{author}{\bibinfo{person}{Yanqiao Zhu}, \bibinfo{person}{Yichen Xu},
  \bibinfo{person}{Feng Yu}, \bibinfo{person}{Qiang Liu}, \bibinfo{person}{Shu
  Wu}, {and} \bibinfo{person}{Liang Wang}.} \bibinfo{year}{2021}\natexlab{}.
\newblock \showarticletitle{{Graph Contrastive Learning with Adaptive
  Augmentation}}. In \bibinfo{booktitle}{\emph{WWW}}.
  \bibinfo{pages}{2069--2080}.
\newblock


\end{thebibliography}

\end{document}